\documentclass[10pt,twocolumn,letterpaper]{article}

\usepackage{wacv}              % To produce the CAMERA-READY version
\usepackage{graphicx}
\usepackage{amsmath}
\usepackage{amssymb}
\usepackage{booktabs}
\usepackage[numbers]{natbib}
\usepackage{microtype}
\usepackage{tabularx}
\newcolumntype{Y}{>{\centering\arraybackslash}X}
\usepackage{algorithm2e}
\usepackage{comment}
\usepackage{multirow}
\usepackage{pifont}% http://ctan.org/pkg/pifont

\usepackage[pagebackref,breaklinks,colorlinks]{hyperref}

\usepackage[capitalize]{cleveref}
\crefname{section}{Sec.}{Secs.}
\Crefname{section}{Section}{Sections}
\Crefname{table}{Table}{Tables}
\crefname{table}{Tab.}{Tabs.}

\def\wacvPaperID{2514} % *** Enter the WACV Paper ID here
\def\confName{WACV}
\def\confYear{2024}

\begin{document}


\title{Semantic Generative Augmentations for Few-Shot Counting}

\author{\begin{tabular}[t]{c@{\extracolsep{1em}}c@{\extracolsep{1em}}c@{\extracolsep{1em}}c}
Perla Doubinsky${}^1$  &
Nicolas Audebert${}^1$
& Michel Crucianu${}^1$ & 
Hervé Le Borgne${}^2$  \\
\end{tabular}
{} \\
 \\
${}^1$ CEDRIC (EA4329), Cnam Paris, France\hfill
${}^2$ Université Paris-Saclay, CEA List, Palaiseau, France 
% ${}^1$ CEDRIC (EA4329), Conservatoire national des arts et métiers, Paris 75003, France   \\
% ${}^2$ Université Paris-Saclay, CEA, List, F-91120, Palaiseau, France 
{} \\
{\tt\small \{perla.doubinsky,nicolas.audebert,michel.crucianu\}@lecnam.net, herve.le-borgne@cea.fr } 
%\\
}
\makeatletter
\let\@oldmaketitle\@maketitle% Store \@maketitle
\renewcommand{\@maketitle}{\@oldmaketitle% Update \@maketitle to insert...
\begin{center}
\vspace{-1em}
\raisebox{20pt}{\includegraphics[width=3.5cm]{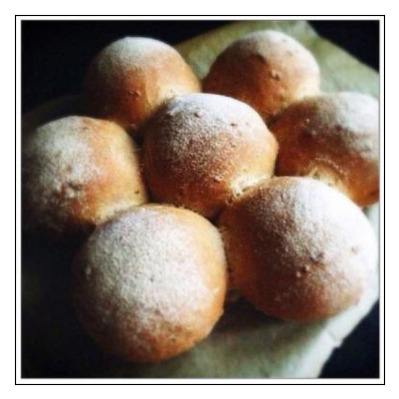}}
\includegraphics[width=10cm]{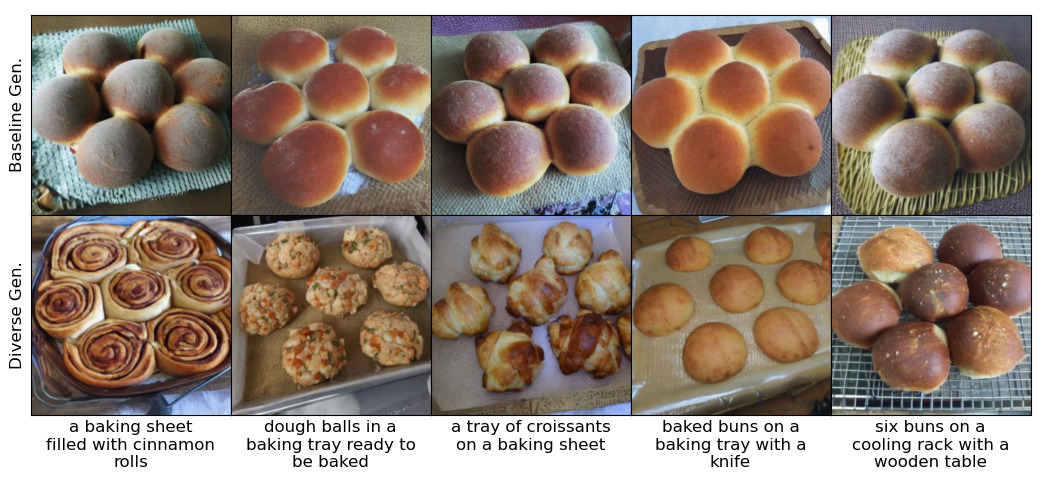}
\captionof{figure}{
Diversified augmentation generation by joint conditioning with density map and semantically similar prompt.
%Diversified augmentations for counting networks with text-and-density conditioned model and caption diversification.
%We generate augmentations for few-shot counting using a diffusion model conditioned on text and \emph{density} maps. For diverse but realistic generations (bottom row), we vary the text input with semantically similar prompt.
}
\bigskip
\label{fig:abstract}
\end{center}
}
\makeatother
\maketitle

\renewcommand{\paragraph}[1]{\noindent\textbf{#1}\quad}

%%%%%%%%% ABSTRACT
\begin{abstract}
%\vspace{-1em}
With the availability of powerful text-to-image diffusion models, recent works have explored the use of synthetic data to improve image classification performances. These works show that it can effectively augment or even replace real data. In this work, we investigate how synthetic data can benefit few-shot class-agnostic counting. This requires to generate images that correspond to a given input number of objects. However, text-to-image models struggle to grasp the notion of count. We propose to rely on a double conditioning of Stable Diffusion with both a prompt and a density map in order to augment a training dataset for few-shot counting. Due to the small dataset size, the fine-tuned model tends to generate images close to the training images. We propose to enhance the diversity of synthesized images by exchanging captions between images thus creating unseen configurations of object types and spatial layout. Our experiments show that our diversified generation strategy significantly improves the counting accuracy of two recent and performing few-shot counting models on FSC147 and CARPK.
\end{abstract}

\section{Introduction}

\begin{figure}
    \centering
    \includegraphics[width=7.5cm]{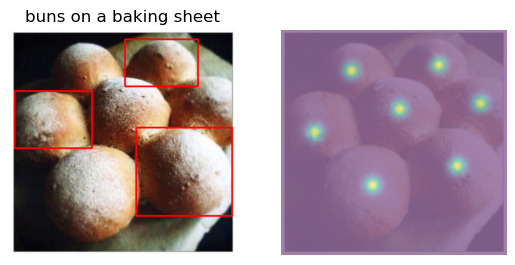}
    \caption{\textbf{Left}: FSC147 image with BLIP2 caption (above) and exemplar boxes (in \textcolor{red}{red}). \textbf{Right}: Ground-truth density map.}
    \label{fig:fsc147_sample}
    \vspace{-1em}
\end{figure}

Counting objects is a task with applications in many domains \eg manufacturing, medicine, monitoring, that involve different types of objects. While earlier works focused on learning specialized networks \cite{arteta2016countingwild,hsieh2017drone,chen2020crowd,bcdata2020}, %\eg cars \cite{hsieh2017drone}, cells \cite{}.
Few-Shot object Counting (FSC) \cite{ranjan2021learning} was recently introduced to train models that can count any object, including from categories outside the training data. Methods tackling FSC rely on \emph{exemplar} objects annotated with bounding boxes (cf.~\cref{fig:fsc147_sample}), in an \emph{extract-then-match} manner \cite{lu2018class}. 
The features of the exemplars and query image are compared using \eg %concatenation \cite{lu2018class}, 
correlation maps \cite{ranjan2021learning,You_2023_WACV} or attention \cite{liu2022countr,djukic2022low}. Matched features are then transformed into a \emph{density} map indicating at each location in the image the density of the objects of interest. The density map is then summed to obtain the predicted count.  

The reference dataset for FSC, namely FSC147 \cite{ranjan2021learning}, contains a limited amount of data (3659 train images) thus bounding the performances of counting networks \cite{ranjan2022vicinal}.
Expanding such a dataset is costly as the annotation process requires pinpointing the center of each object present in a query image, with a potentially high number of occurrences. To overcome the small dataset size, Ranjan \emph{et al.}~\cite{ranjan2022vicinal} augment FSC147 using a GAN to diversify the image styles.
Diffusion models have now surpassed GANs owing to their training stability and lower sensitivity to mode collapse. These models produce more effective and diverse augmentations \cite{he2022synthetic,trabucco2023effective,shiparddiversity}.
Recent works mostly aim at augmenting classification datasets \eg ImageNet~\cite{imagenet_cvpr09}, where augmentations are generated by prompting the models with the image labels. This fails to produce satisfying images for counting datasets as text-to-image models struggle to generate the correct number of objects \cite{petsiuk2022human}. Some works tackle improving compositionality in vision-language models \cite{paiss2023teaching,lee2023aligning,phung2023grounded} but are limited to small numbers of objects. Other works add more control to pre-trained text-to-image models \cite{zhang2023adding,lhhuang2023composer,mou2023t2i}.

To tackle few-shot counting, we propose to synthesize unseen data with Stable Diffusion conditioned by both a textual prompt and a density map. We thus build an augmented FSC dataset that is used to train a deep counting network. The double conditioning, implemented with ControlNet \cite{zhang2023adding}, allows us to generate novel synthetic images with a precise control, preserving the ground truth for the counting task. It deals well with large numbers of objects, while current methods fail in such cases~\cite{lee2023aligning,phung2023grounded}. To increase the diversity of the augmented training set, we swap %some prompts (or alternatively some density maps)%
image descriptions between the $n$ available training samples, leading to $\frac{n(n-1)}{2}$ novel couples, each being the source of several possible synthetic images. However, we show that some combinations do not make sense and lead to poor quality samples. Therefore, we only select plausible pairs, resulting in improved augmentation quality.
We evaluate our approach on two class-agnostic counting networks, namely SAFECount \cite{You_2023_WACV} and CounTR \cite{liu2022countr}. We show that it significantly improves the performances on the benchmark dataset FSC147 \cite{radford2021learning} and allow for a better generalization on the CARPK dataset \cite{hsieh2017drone}. 

\section{Related work}

\paragraph{Learning with Generated Data}
Improvements in image synthesis using generative models have sparked great interest in generating fake images to train deep neural networks. GANs were the first popular models to synthesize data for image classification \cite{antoniou2017data,besnier2020dataset,jahanian2021generative}, crowd counting \cite{wang2019learning} and image segmentation \cite{zhang2021datasetGAN}. Nowadays, diffusion models such as DDPM \cite{ho2020denoising} or Latent Diffusion \cite{rombach2022high} seem to outperform GANs, demonstrating more stable training, better coverage of the training distribution and higher image quality.
The availability of powerful text-conditioned diffusion models \cite{rombach2022high, nichol2021glide,ramesh2022hierarchical,saharia2022photorealistic} has led to many works exploring how to leverage synthetic data for computer vision, \eg image classification in low-data regime \cite{he2022synthetic}, zero/few-shot learning \cite{trabucco2023effective,shiparddiversity}, ImageNet classification \cite{bansal2023leaving,sariyildiz2023fake,azizi2023synthetic} and self-supervised learning \cite{tian2023stablerep}. %, bakhtiarnia2023promptmix}.
These works focus on how to reduce domain gap \cite{he2022synthetic}, improve the prompts using \eg text-to-sentence model \cite{he2022synthetic} or WordNet \cite{sariyildiz2023fake} and increase diversity by optimizing the guidance scale \cite{sariyildiz2023fake,shiparddiversity,azizi2023synthetic}. 
%how to mix synthetic data and real data.
This body of literature consistently demonstrates how generated data %has the potential to 
allow deep networks to learn more robust representations and improve generalization for image classification.
%and crowd counting in PromptMix \cite{bakhtiarnia2023promptmix}.
%PromptMix \cite{bakhtiarnia2023promptmix} uses synthetic data to improve crowd counting. 
We focus more specifically on few-shot class-agnostic object counting. Compared to image classification, this task involves small datasets and local spatial understanding, as objects can be small and follow complex layouts. The generated data needs a level of compositionality that current generative models, including diffusion models, struggle to achieve.
To bring the power of synthetic data to counting, we propose to condition diffusion models not only on text prompts but also on counting density maps to generate images with the correct number of objects in the desired spatial configuration. We exploit this double control to generate diversified unseen data by prompting the model with novel combinations of the controls. 

\paragraph{Few-shot Object Counting}
The goal of few-shot class-agnostic object counting is to count how many instances of objects of \emph{any} arbitrary category there are in a given image, by leveraging only \emph{a few} exemplars of the category of interest.
This was initially formulated as %a 
matching %problem between 
exemplars and %query
image patch features~\cite{lu2018class}.
FSC147~\cite{ranjan2021learning} was later put forward as the main dataset for this task, with an open set train and test split to evaluate generalization to unseen object categories. Its authors introduced FamNet, a deep net trained to infer density maps from feature similarities. In the same lineage, BMNet \cite{shi2022represent} refines the similarity map by learning the similarity metric jointly with the counting network. In SAFECount \cite{You_2023_WACV}, the similarities are used to fuse exemplars features into the query image features. The density map is then predicted from the enhanced features. Other works \eg CounTR \cite{liu2022countr} and LOCA \cite{djukic2022low} focus on improving the feature representations using a Transformer backbone as the visual encoder and injecting information about the exemplars' shape in the network~\cite{djukic2022low}.
%Recent works now also consider to use no exemplars %, namely zero-shot counting, 
%by leveraging CLIP \cite{jiang2023clip}.
The closest comparison to our work is the Vicinal Couting Network from Rajan \emph{et al.} \cite{ranjan2022vicinal}. It augments FSC147 with generated data by training a conditional GAN jointly with the counting network, producing augmentations that preserve the image content while modifying its visual appearance. While outperformed by later models, it introduced the idea that well-chosen augmentations can significantly boost counting accuracy. In this work, we leverage large pre-trained text-to-image diffusion models to produce diverse augmentations that not only alter the appearance, but are also able to change the content, to synthesize augmentations with a variety of object semantics.% and the spatial layout. 
%generalization ability t2i models.

\begin{figure}
    \centering
    \includegraphics[width=8cm]{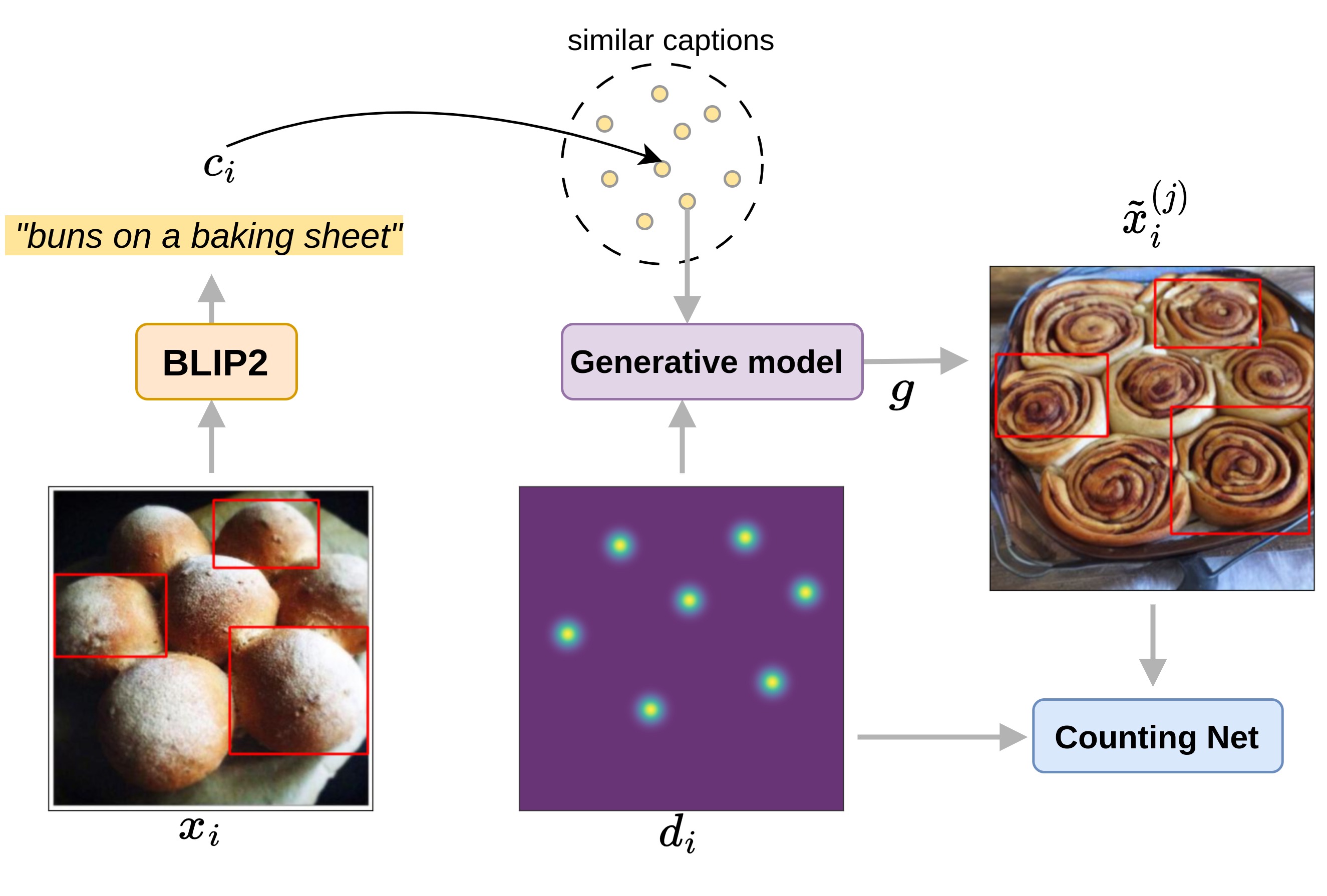}
    \vspace*{-10pt}
    \caption{Overview of our approach. We condition a pre-trained diffusion model on both text prompts and density maps and perform swaps with similar captions. The density and original exemplars boxes are used as ground-truth for the generated augmentation.
    }
    \label{fig:overview_method}
    \vspace{-1em}
\end{figure}

\section{Latent diffusion}
\label{sec:background}

Rombach \emph{et al.}~\cite{rombach2022high} introduced Latent Diffusion Models (LDMs) that are diffusion models applied in the latent space of generative models such as VQGAN \cite{esser2020taming}. These models operate in a compressed latent space thus reducing training and inference time. For more controlled generation, LDMs can be conditioned with \eg text, images or semantic maps. 
%via cross attention.
%Given the condition $c_t$, the objective of LDMs is: %(\textcolor{blue}{U-Net, cross attention})
%\begin{align}
%    \mathcal{L} = ||\epsilon_t - \epsilon_{\theta}(z_t, t, c_t)||^2 
%\end{align}
%where $z_t$ is the latent representation of $x_t$.
Stable Diffusion is a popular open-source LDM, trained on the large-scale LAION-2B dataset \cite{schuhmann2022laion} and enabling high-quality text-to-image generation. Prompt-based control heavily relies on the capabilities of the underlying text encoder, typically CLIP \cite{radford2021learning} that is known to poorly integrate compositional concepts such as counting \cite{paiss2023teaching}. %(\textcolor{blue}{link})
%improve
%Thus, recent works focus on adapting these models by relying on additional control signals
%Recent works focus on allowing fine-grained control while retaining the expressiveness of pre-trained models 
%\cite{zhang2023adding,lhhuang2023composer,mou2023t2i}. 

\paragraph{ControlNet}
%give more details
Recent works have added ancillary control signals to LDMs. Such a popular approach is ControlNet~\cite{zhang2023adding}, which extends available %existing
pre-trained %latent 
diffusion models by creating a trainable copy of the original network with an additional control input such as a semantic, edge or depth map. This trainable copy is linked to a locked copy %through zero convolution 
to preserve the capacities of the original model. %when training with a small dataset. 
The training objective is an LDM objective similar to \cite{rombach2022high}:%\cref{denoising_objective}:
\begin{align}
    \mathcal{L} = ||\epsilon_t - \epsilon_{\theta}(z_t, t, c_t, c_f)||^2 
\end{align}
where $\epsilon_t \sim \mathcal{N}(0, 1)$, $\epsilon_{\theta}(,)$ is a model of the diffusion noise, $z_t$ is the latent representation of $x_t$, $c_t$ the text prompt and $c_f$ the task-specific control.

\begin{figure*}[t]
    \begin{center}
    \begin{subfigure}{0.15\textwidth}
        \caption*{Real image}
        \centering
        %\vspace{-3cm}
        \raisebox{50pt}{\includegraphics[width=2.5cm]{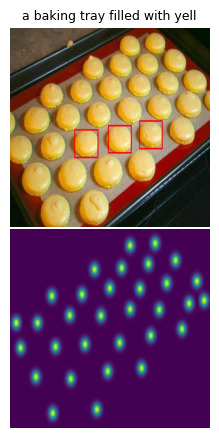}}
%        \caption{Real image}
%        \caption{}
    \end{subfigure}  
    \hspace{-1.5cm}
    \begin{subfigure}{0.84\textwidth}               \caption*{Synthetic augmentations}
        \centering
        \includegraphics[width=12cm]{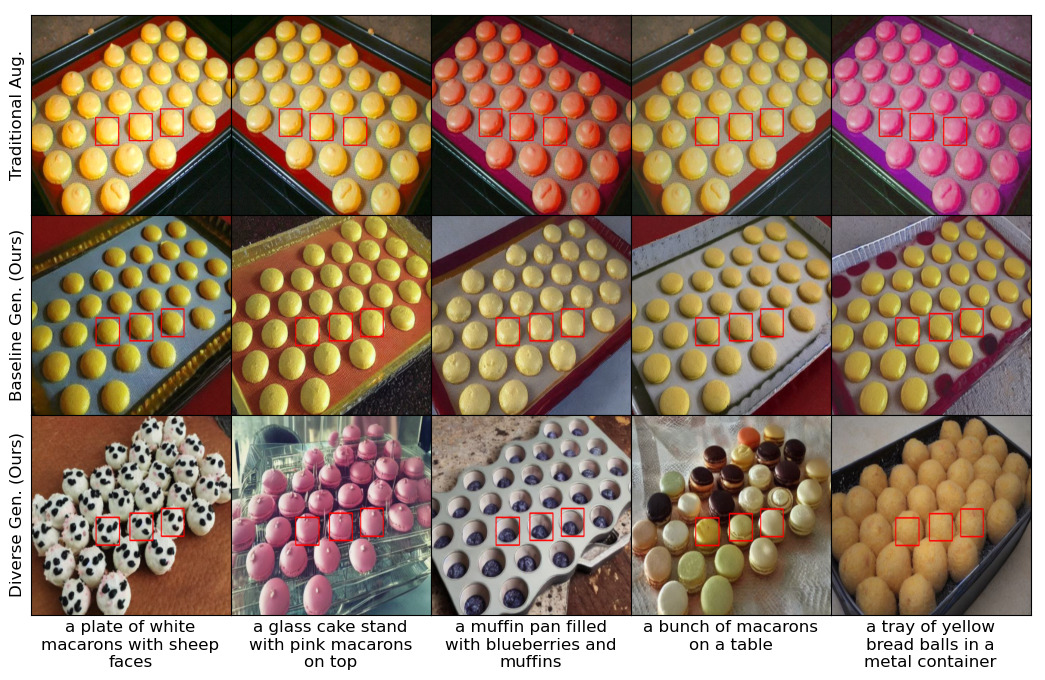}
%        \caption{Synthetic augmentations}
    \end{subfigure}
    \begin{subfigure}{0.15\textwidth}
        \centering
        \raisebox{14pt}{\includegraphics[width=2.5cm]{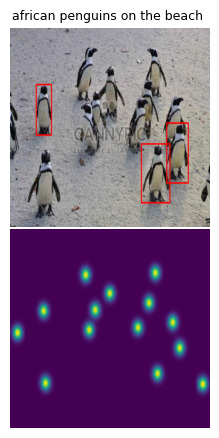}}
        %\caption{Real image}
    \end{subfigure}  
    \hspace{-1.5cm}
    \begin{subfigure}{0.84\textwidth}
       \centering
        \includegraphics[width=12cm]{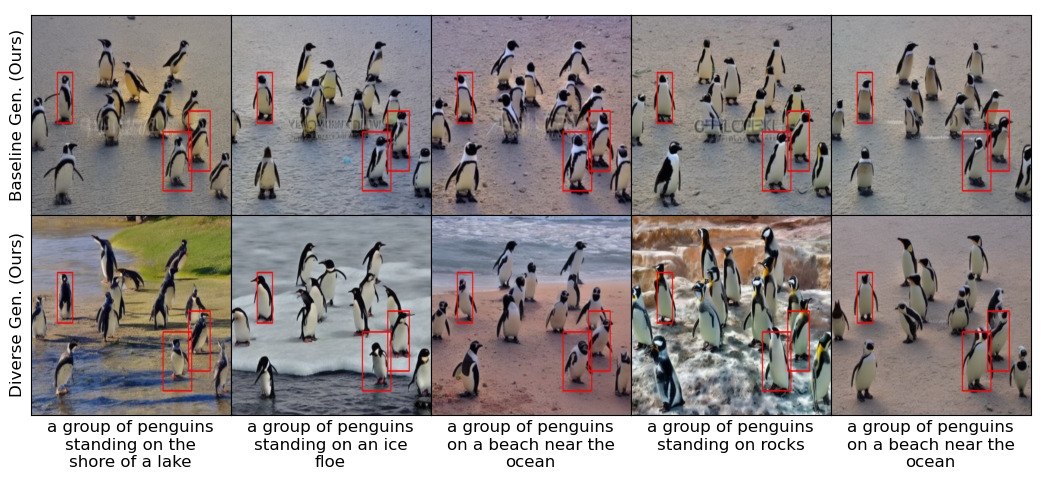}
        %\caption{Synthetic augmentations}
    \end{subfigure}
    \begin{subfigure}{0.15\textwidth}
        \centering
        \raisebox{14pt}{\includegraphics[width=2.5cm]{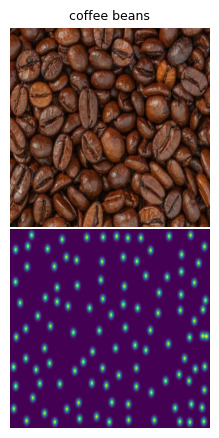}}
        %\caption{Real image}
        %\caption*{Real image}
    \end{subfigure}     
    \hspace{-1.5cm}
    \begin{subfigure}{0.84\textwidth}
        \centering
        \includegraphics[width=12cm]{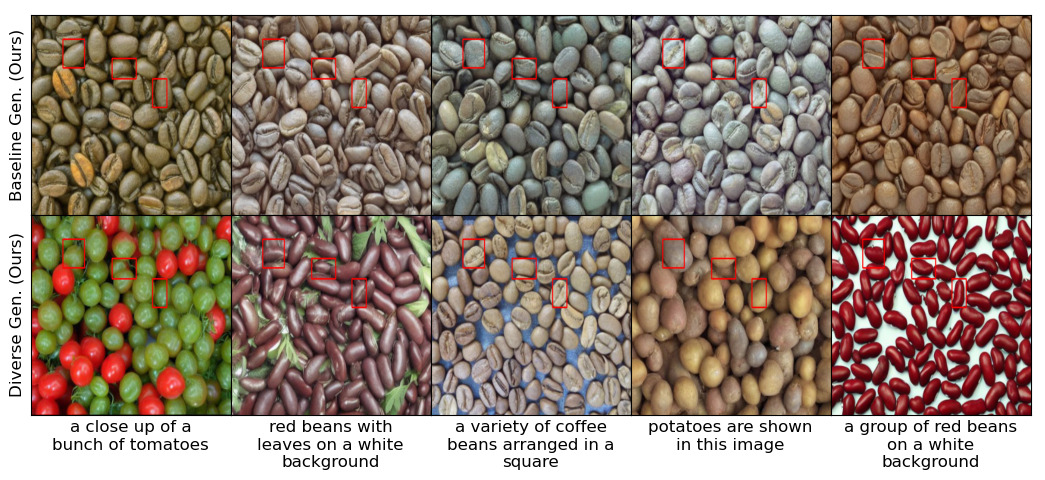}
        %\caption*{Synthetic augmentations}
    \end{subfigure}
    \end{center}
    \vspace{-10pt}\caption{Qualitative results for the Baseline vs. Diverse augmentations. At the bottom of each diverse sample we show the caption used to generate the image. Our strategy allows to diversify the type of objects and/or the background.
    }
    \label{fig:visu_aug}
    \vspace{-1em}
\end{figure*}

\section{Semantic Generative Augmentations}
%\section{Generative Augmentations for FSC}%change 
%We first formally define few-shot object counting, before giving an overview of our generative model and presenting our synthetic data augmentation strategy.% Then, we give an overview of the generative model. Finally, we present our strategy to augment the data and diversify the augmentations. %change

\subsection{Few-Shot Counting}
\label{subsec:overview}

The goal of few-shot class-agnostic counting is to learn to count objects regardless of their category. To achieve this, the query image $x\in \mathbb{R}^{H\times W \times 3}$ is annotated with $n \in \{0,1,2,3...\}$ \emph{exemplar} boxes of coordinates $b\in \mathbb{R}^{4}$. %taken from the query or external images.
The counting network takes as input both the query image and the set of $n$ boxes. It predicts a density map %\cite{NIPS2010_fe73f687}
$d \in \mathbb{R}^{H \times W}$ of same size as the image. As shown in \cref{fig:fsc147_sample}, this ground-truth density map has zero values where there are no objects, and a Gaussian kernel of fixed variance at the center of every object. % is a spatial map with 1's at the objects' centers and 0's at locations where there are no objects. %Gaussian
The final count is obtained by summing across all positions of the density map. % by taking the sum over the density map.
The model is typically trained with an $L_2$ loss between the predicted and ground-truth densities. %There are usually three datasets $\mathcal{D}_\text{train}$, $\mathcal{D}_\text{val}$, $\mathcal{D}_\text{test}$ comprising objects of disjoints categories. The goal is to learn a counting network on $\mathcal{D}_\text{train}$ able to count the unseen objects in $\mathcal{D}_\text{val}$ and $\mathcal{D}_\text{test}$. 
To evaluate class-agnostic models, object categories from the test set %$\mathcal{D}_\text{test}$ 
are disjoint from those in the validation %$\mathcal{D}_\text{val}$ 
and train sets. %$\mathcal{D}_\text{train}$. 
This open set evaluation allows us to measure the network's ability to %such that the counting network must generalize its ability to 
count objects from \textit{unseen} categories.

\subsection{Text-and-Density Guided Augmentations}
%improve text
To synthesize new images that can effectively augment a few-shot counting dataset, we need to have control over the number of objects and how they are laid out.
%reformulate
Indeed, we need to ensure that we know the density maps of the synthetic samples so that they can be used to train the model.
In addition, being able to control object type and spatial configuration also constitutes a lever to diversify the dataset by generating new combinations of categories and densities.
%It allows circumventing the labeling of the generated data and also constitutes a lever to diversify it. 
As few-shot counting datasets are generally limited in size, we take advantage of available pre-trained diffusion models to synthesize diversified augmentations of the training samples, reducing overfitting and improving generalization.
However, large pre-trained generative models such as Stable Diffusion are usually conditioned through textual prompts. %useful to vary semantics of objects 

To finetune these models, we first have to pair textual captions to the training images. We obtain diverse and descriptive captions using an off-the-shelf captioning model, \eg BLIP2 \cite{li2023blip}. This produces richer captions than plain object categories such as ``a photo of \{class\}``.
%which may not always be available
However, two shortcomings remain. First, generated captions may not contain any information about the number or arrangement of the objects. Second, text-conditioned LDMs poorly respect prompts regarding compositional constraints. Even adding this information in the caption does not guarantee that generated images would follow them. This is especially problematic as the correctness of the layout is a prerequisite to generate images for which we know the ground-truth.
Therefore, we further condition the generative model directly on the density maps as an additional input, using the ControlNet fine-tuning strategy.
To summarize, our generative model is now conditioned on a text prompt, obtained by an automated captioning of the training image, and its ground truth density map to enforce the spatial layout of the objects.
This allows us to synthesize new samples that augment the original image, while keeping the ground truth intact, making the augmentation amenable to supervised learning.

\subsection{Diversity-Enhanced Augmentations}

To formalize the augmentation process, let $\mathcal{D}_{\text{train}} =\{x_i, b_i, d_i\}_{i=1}^N$ be an annotated counting dataset, with $x_i$ an image, $b_i$ the exemplar bounding boxes for each image, and $d_i$ its ground-truth density map. Let $\mathcal{C}=\{c_i\}_{i=1}^N$ be the set of corresponding captions. For each image $x_i$, we aim at generating $M$ augmentations using our text-density conditional generative model $g(d_i, c_i)$.
%We introduce,...

\paragraph{Baseline} We sample augmentations from the LDM by taking advantage of the non-deterministic \emph{reverse} diffusion process and the expressiveness of the pre-trained model. For an image $x_i$ we produce $M$ augmentations $\Tilde{x}_i^{(j)}$ that share its caption and density map:
\begin{align}
    \Tilde{x}_i^{(j)} = g(d_i, c_i), \quad j=1,...,M
\end{align}
These augmentations preserve both the number and layout of objects -- because of the density conditioning -- and the semantics \eg object category and type of background -- because of the text prompt. This already augments the number of samples available for training.

\paragraph{Diverse} 
We can however go further and \emph{diversify} the augmentations by altering either the text description or the spatial organisation of the objects. To do so, we take advantage of dual conditioning on both densities and captions. We mix the two sets to create new combinations (density map, caption), producing augmentations that are semantically and geometrically more diverse than the original dataset.
Yet, this mixing of the conditionings should be done carefully, to avoid low quality augmentations. Indeed, not all combinations make sense, \eg ``a herd of cows'' and ``a pearl necklace'' exhibit very different spatial layouts.
To prompt the generative model with realistic (\emph{density}, \emph{text}) pairs, we rely on caption similarity to find new associations between images that share some semantics, \eg ``cows'' and ``bisons''.

\label{subsec:swap_caption}
We swap captions at random between pairs of \emph{compatible} images. Two images are said to be compatible if their captions are more similar than some threshold $t_c$, \ie:
$$\text{sim}(c_i, c_k) = \frac{\Psi(c_i)^{\top}\Psi(c_k)}{||\Psi(c_i)||_2 ||\Psi(c_k)||_2} > t_c$$
where $\Psi$ is a suitable text encoder.
We then sample new images using the initial density map, but replacing the original caption with the caption $c_k \in \mathcal{C}$ from a compatible training observation chosen at random:
\begin{align} 
    \Tilde{x}_i^{(j)} = g(d_i, c_k), \quad j=1,...,M
\end{align}
This process results in more diverse augmentations compared to the baseline and alters more the images than traditional augmentations (color jitter, crops, etc.), as shown in \cref{fig:visu_aug}.

\paragraph{Synthetic and Diverse Balance}
We follow the training strategy from Trabucco \emph{et al.}~\cite{trabucco2023effective}, where the synthetic augmentations are used as a regular data augmentation with a probability $p_0$ when training the counting model. As a way to balance baseline and diversified augmentations, we set a probability $p_c$ that defines the fraction of the $M$ augmentations that use a swapped caption instead of the original one. Typically, $p_c = 0.5$ means that 50\% of the generated augmentations employ the original (caption, density) pair and that the remaining 50\% use new (caption, density) combinations.
For each augmentation, we keep the density used to condition the image generation and the original exemplar boxes as ground truth to train the model.\footnote{Note that if the caption changes the object category, bounding boxes for the exemplars might not be accurate anymore (\eg ``pens'' are narrow and elongated, while ``erasers'' are closer to squares, see \cref{sec:limitation}.}

\begin{table*}[ht!]
\begin{subtable}{0.6\linewidth}
    %\centering
    \caption{SAFECount \cite{You_2023_WACV}}
    \begin{tabularx}{\linewidth}{lYYYY|
%    |YYYY
    }
        \toprule
        %& \multicolumn{4}{c|}{3-shot} 
        %& \multicolumn{4}{c}{1-shot} 
        %\\
        & \multicolumn{2}{c}{Val} & \multicolumn{2}{c|}{Test} 
        %& \multicolumn{2}{c}{Val} & \multicolumn{2}{c}{Test} 
        \\
        \midrule
        %\cmidrule{2-9}
        %\cmidrule{0-5}
        & MAE & RMSE & MAE & RMSE 
        %& MAE & RMSE & MAE & RMSE 
        \\ 
        \midrule
        Traditional Augmentation$^{*,\dagger}$ & 13.95 & 51.73 & 13.73  & 91.85 
        %&  &  &  &  
        \\
        \midrule
        + Real Guidance~\cite{he2022synthetic} & 14.94 & 53.09 & 13.48 & \textbf{80.69} %&&&&
        \\
        + Baseline Generation (Ours) & 13.30 & 49.38 & 13.22  & 92.47 
        %&&&&
        \\
        + Diverse Generation (Ours) & \textbf{12.59} & \textbf{44.95} & \textbf{12.74} & 89.90 
        %&  &  & &  
        \\
        \midrule
        \midrule
        Traditional Augmentation (reported) & 15.28 & 47.5 & 14.25 & 85.54 \\
        \bottomrule
    \end{tabularx}
    %\caption{Quantitative results on FSC147 for SAFECount \cite{You_2023_WACV}.}
    \label{tab:safecount_fsc147}
    \end{subtable}
%\end{table}
\begin{subtable}{0.4\linewidth}
%\begin{table}[t!]
    %\centering
        \caption{CounTR \cite{liu2022countr}}
    \begin{tabularx}{\linewidth}{lYYYY %|YYYY
    }
        \toprule
        %& \multicolumn{4}{c|}{3-shot} 
        %& \multicolumn{4}{c}{0-shot} 
        %\\
        & \multicolumn{2}{c}{Val} & \multicolumn{2}{c}{Test}
        %& \multicolumn{2}{c}{Val} & \multicolumn{2}{c}{Test} 
        \\
        %\cmidrule{2-9}
        %\cmidrule{0-5}
        \midrule
        & MAE & RMSE & MAE & RMSE 
        %&MAE & RMSE & MAE & RMSE 
        \\ 
        \midrule
        %CounTR$^{*}$ 
        & 14.25 & 50.15 & 13.13 & 88.21
        %&  &  &  &
        \\ 
        \midrule 
        %Real Guidance 
        & 15.37 & 49.47 & 13.37 & 96.44 
        %& & & &
        \\
        %Baseline (Ours) 
        & 12.60 & 43.53 & 11.83 & 87.97
        %& & & &
        \\
        %Diverse (Ours) 
        & \textbf{12.31} & \textbf{41.65} & \textbf{11.32} & \textbf{77.50} 
        %&  &  &  & 
        \\
        \midrule 
        \midrule
        & 13.13 & 49.83 & 11.95 & 91.23 \\
        \bottomrule
    \end{tabularx}
    \label{tab:countr_fsc147}
    \end{subtable}
    \caption{Quantitative results on FSC147. (*) Traditional augmentations include color jitter, random cropping. ($\dagger$) \cite{You_2023_WACV} and \cite{liu2022countr} are reproduced, details are provided in the supplementary material.}
    \label{tab:quant_results_fsc147}
\end{table*}

\section{Experiments}
%add CARPK if included
\subsection{Datasets and metrics}
\paragraph{FSC147} FSC147 \cite{ranjan2021learning} is a 3-shot counting dataset with 147 object categories. It is the \emph{de facto standard} of class-agnostic counting benchmarking. 89 categories are used for the training set, 29 are included in the validation set the remaining 29 constitute the test set. Note that the categories from the three sets are completely disjoint. In total, the dataset contains 6135 images, from which 3659 are used for training. The number of objects in the images varies from 7 to 3731 with an average of 56. Every image is annotated with 3 exemplar bounding boxes and an object density map.

\paragraph{CARPK} 
%To measure the generalization ability of our model
CARPK \cite{hsieh2017drone} is a class-specific dataset for counting cars in parking lots based on overhead imagery. The dataset contains 1448 images from a UAV in 4 parking lots: 3 for training and the last one for testing. There are 5 exemplar objects in total that are randomly extracted from the training set and employed during both training and testing. Following \cite{You_2023_WACV,liu2022countr}, we evaluate on CARPK the generalization ability of our class-agnostic models on a new dataset.

\paragraph{Metrics} %To evaluate counting networks, 
We follow the standard evaluation of the counting accuracy through the Mean Absolute Error (MAE) and Root Mean Squared Error (RMSE). 
%For a ground-truth count $y_i$ and  a predicted count $\hat{y}_i$, we define:
%\begin{align}
%\text{MAE} = \frac{1}{N}\sum_{i=1}^N|y_i - \hat{y}_i|, \quad \text{RMSE} = \sqrt{\frac{1}{N}\sum_{i=1}^N(y_i - \hat{y}_i)^2}  
%\end{align}

\subsection{Implementation details}

\paragraph{Augmentation}
We train ControlNet on the training images and density maps from FSC147. Text prompts are obtained by captioning the images with BLIP2~\cite{li2023blip}. The underlying pre-trained diffusion model is Stable Diffusion $v1.5$ trained on LAION 2B. We use 
%a learning rate of $1e^{-5}$ and batch size of $4$, following 
the default settings and train for 350 epochs. After training, we employ a guidance scale of $2.0$ and $20$ denoising steps to generate an image.
For each augmentation strategy, we generate $M=10$ augmentations per training sample unless specified otherwise. We swap the original caption with another one with a probability $p_c=0.5$. Compatible captions to swap with are obtained by extracting caption features with the BLIP2 text encoder and filtering the captions with a similarity higher than $t_c=0.7$.

\paragraph{Counting networks}
We demonstrate the effectiveness of our augmentation strategies on two state-of-the-art counting networks: SAFECount \cite{You_2023_WACV} and CounTR \cite{liu2022countr}. SAFECount is a CNN while CounTR is Transformer-based. In CounTR, training is done in two phases. First, the network is pre-trained using a self-supervised masked auto-encoder \cite{he2022masked}, then it is fine-tuned in a supervised fashion with the usual $L_2$ loss on the densities. For CounTR, we employ the pre-trained model released by the authors and only retrain the fine-tuning phase. We use the hyperparameters reported in the original papers to train both networks, except training is 100 epochs longer to account for the higher number of training images. During training, we replace an image $x_i$ with one of its augmentations $\Tilde{x}_i^{(j)}$ with probability $p_0=0.5$. This balances the ratio of real vs.\ synthetic data in a single batch. We also employ traditional data augmentation strategies \eg flips, color jitter, random cropping, that are applied to every image, both real and synthetic, as done in the original models. Both networks are trained and evaluated in the 3-shot setting. All results are averages over two runs.

\subsection{Few-Shot Counting on FSC147}

\paragraph{Comparison with Traditional Augmentation}
We report in \cref{tab:quant_results_fsc147} the improvement in counting accuracy on FSC417 with our augmentation strategies when training SAFECount and CounTR. Consistent with the literature on synthetic data augmentation, baseline augmentations %due to the non-deterministic reverse diffusion process (cf.~\cref{subsec:baseline_augmentation}) %slightly 
improve the results for both networks: MAE decreases by respectively 5\% and 10\% for SAFECount and CounTR on the val set. Nonetheless, diversifying the augmentations allows us to reduce the MAE even further, by 10\% and 11\% on the same val set and by 7\% (SAFECount) and 13\% (CountTR) on the test set.
We attribute this to ControlNet overfitting the training data due to the small dataset size. The low guidance employed to generate the images (2.0) aims at promoting diversity \cite{sariyildiz2023fake} but, as shown in \cref{fig:visu_aug} (Baseline Gen.), the generated images remain close to the original image in terms of visual appearance of the objects and background. %are similar to this image. %size, orientation of the objects.
%Despite some overfitting on the training combinations of density maps and text prompts,
However, ControlNet generalizes to different captions. In \cref{fig:visu_aug} (Diverse Gen.), we observe that swapping captions allows us to create more diverse data, altering the size and texture of objects and their background. Such features cannot be altered with traditional data augmentation.% and this type of data augmentation is thus complementary to ours.
When mixing baseline and diverse augmentations, the performances for both networks improve significantly with respect to the model without synthetic augmentation, or with naive augmentations only.

\begin{figure}[ht]
    \centering
    \includegraphics[width=8cm]{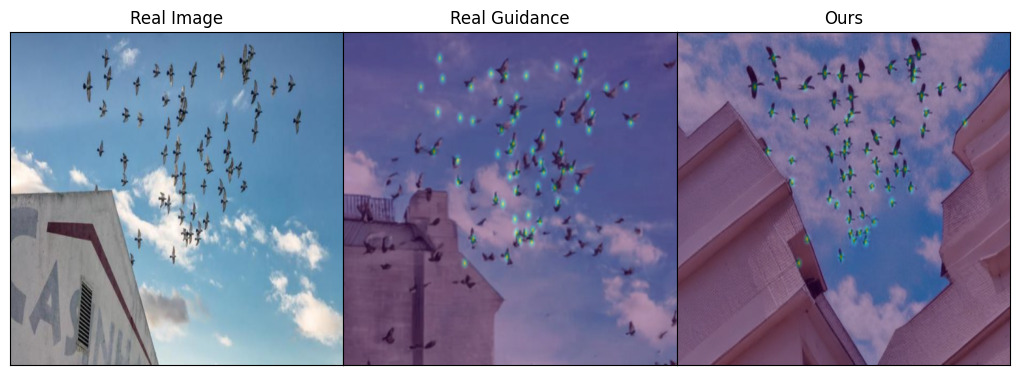}
    \includegraphics[width=8.1cm]
    {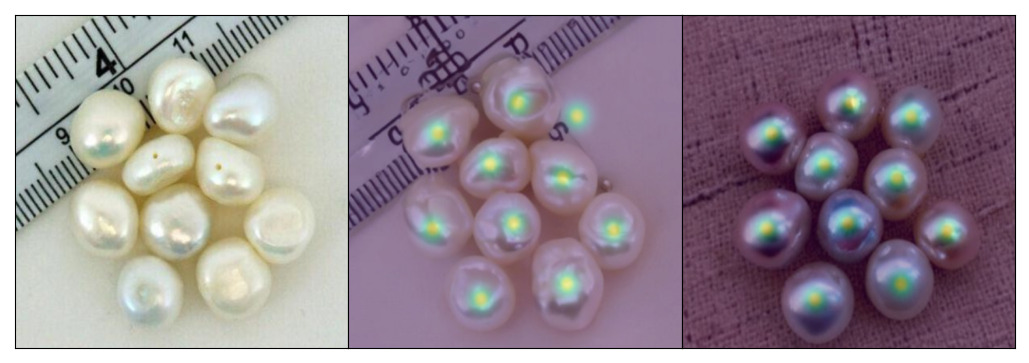}
    \caption{Qualitative comparison with Real Guidance \cite{he2022synthetic}. Our augmentations preserve the layout while creating more diverse backgrounds. Ground-truth density maps overlap with the generated images (last 2 columns).}
    \label{fig:qualitative_comparison_rg}
    \vspace{-1em}
\end{figure}

%Real Guidance comparison
\paragraph{Comparison with Real Guidance}
We compare our approach with Real Guidance, an augmentation strategy for image classification by He~\emph{et al.}~\cite{he2022synthetic}. Augmentations are generated by prompting a pre-trained text-to-image diffusion model with the image classes. To reduce the domain gap, the synthetic images are generated from the real images with added noise as proposed in SDEdit \cite{meng2021sdedit}.
%For fair comparison, 
%We employ Stable Diffusion and the object categories as textual prompts.
%The results presented in 
\cref{tab:quant_results_fsc147} shows that our augmentation strategy outperforms Real Guidance\footnote{Except on test RSME with SAFECount, where Real Guidance performs better, due to two outlier test images with more than 2500 objects that dominate the average error (see supplementary material).}. Starting from the real image with added noise is generally insufficient to preserve the number of objects and their positions (\cref{fig:qualitative_comparison_rg}, $2^{nd} col.$). It shows that the density map conditioning ensures the preservation of object positions and number without requiring to start from the real image, which can limit the diversity of the generated images.
%as shown in Fig...

\subsection{Ablation Study}
We study here the influence of the hyperparameters of our approach: impact of the caption similarity threshold, $t_c$, %used to exchange only among plausible captions w.r.t.\ a given density map,
influence of introducing diversified augmentations by varying the $p_c$ parameter,
the effect of the number of augmentations $M$,
and of the ratio of synthetic samples during training, $p_0$. All ablations are conducted on SAFECount trained for 200 epochs to reduce training time.

\begin{figure}[t]
    \centering
   \includegraphics[width=2cm]{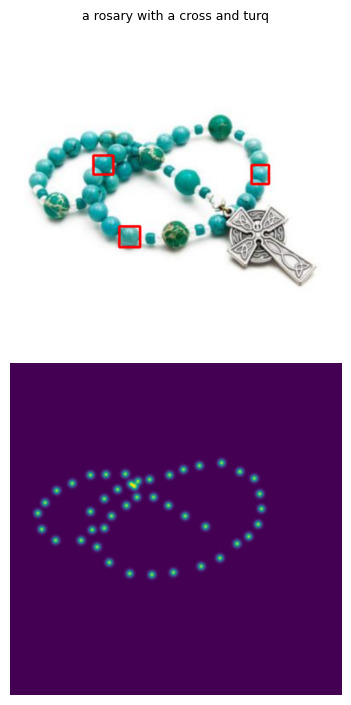}    \includegraphics[width=6cm]{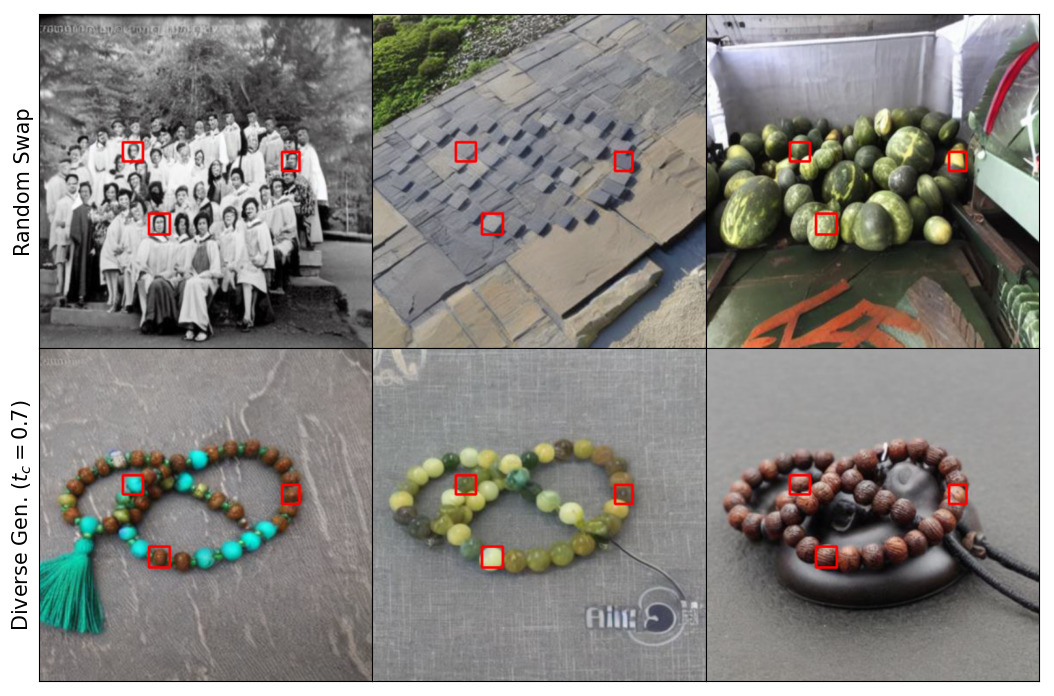}
    \caption{Caption swap at random (top) vs. similarity-based swap (bottom, $t_c=0.7$). Random swapping results in a mismatch between the layout and the semantics.
    }
    \label{fig:random_vs_caption_sim}
    \vspace{-1em}
\end{figure}

\begin{figure}[t]
    \centering
    \begin{subfigure}{\linewidth}
    \includegraphics[width=8cm]{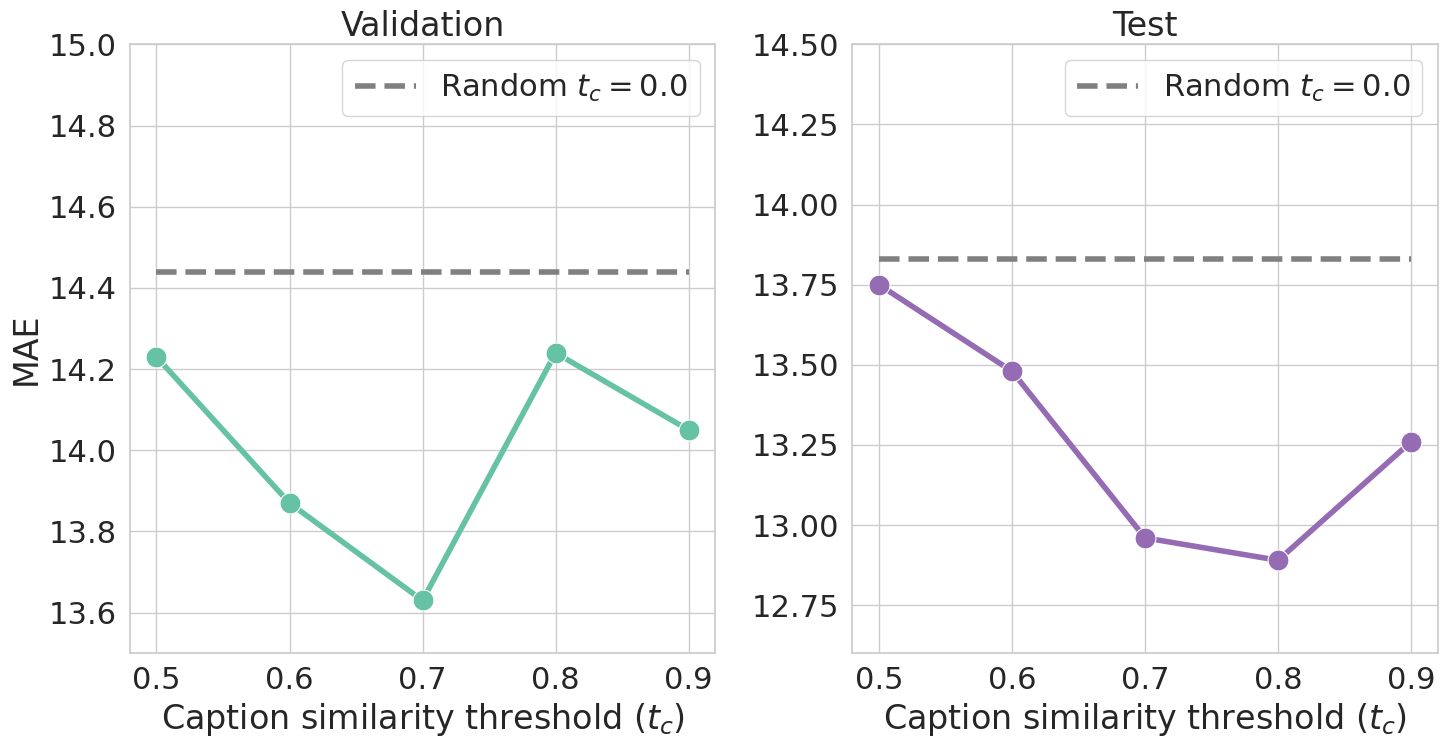}
    \caption{Validation and test MAE on FSC147 w.r.t.\ $t_c$ (SAFECount).}
    \label{fig:ablation_tc}
    \end{subfigure}
    \begin{subfigure}{\linewidth}
    \includegraphics[width=8cm]{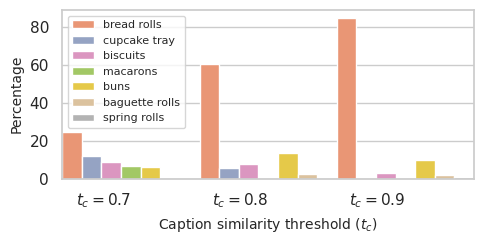}
    \caption{Distribution of object categories in the set of captions to swap from w.r.t\ $t_c$ for a sample of category ``bread rolls''. Lower thresholds result in more diverse augmentations, while objects still belong to similar classes.}
    \label{fig:categories_tc}
    \end{subfigure}
    \caption{Impact of caption similarity threshold $t_c$ on SAFECount.}
\end{figure}

\paragraph{Caption Similarity Threshold} We swap captions based on caption similarity to form novel but plausible (\emph{density}, \emph{text}) combinations. As shown in \cref{fig:random_vs_caption_sim}, associating a completely unrelated caption to a given density map results in generated images that do not correspond to the input density map or are of poor quality, as it is harder for the model to generalize. 
In \cref{fig:ablation_tc} we evaluate different similarity thresholds to the naive approach where all captions can be swapped freely at random ($t_c=0.0$). 
%We find that from $t_c\geq 0.6$ the performances are higher than with random swap. 
The performances are improved compared to random swaps with all thresholds between $0.5$ and $0.9$.
However, there is a quality-diversity tradeoff shown in \cref{fig:categories_tc}. Setting the threshold too high ($t_c=0.8, 0.9$) swap captions between images of objects belonging to the same category, thus limiting diversity. With a lower threshold, \eg $t_c=0.7$, new captions can also belong to objects from different similar categories, \eg swapping ``bread rolls'' and ``macarons''.

\begin{figure}[t]
    \centering
    \includegraphics[width=8cm]{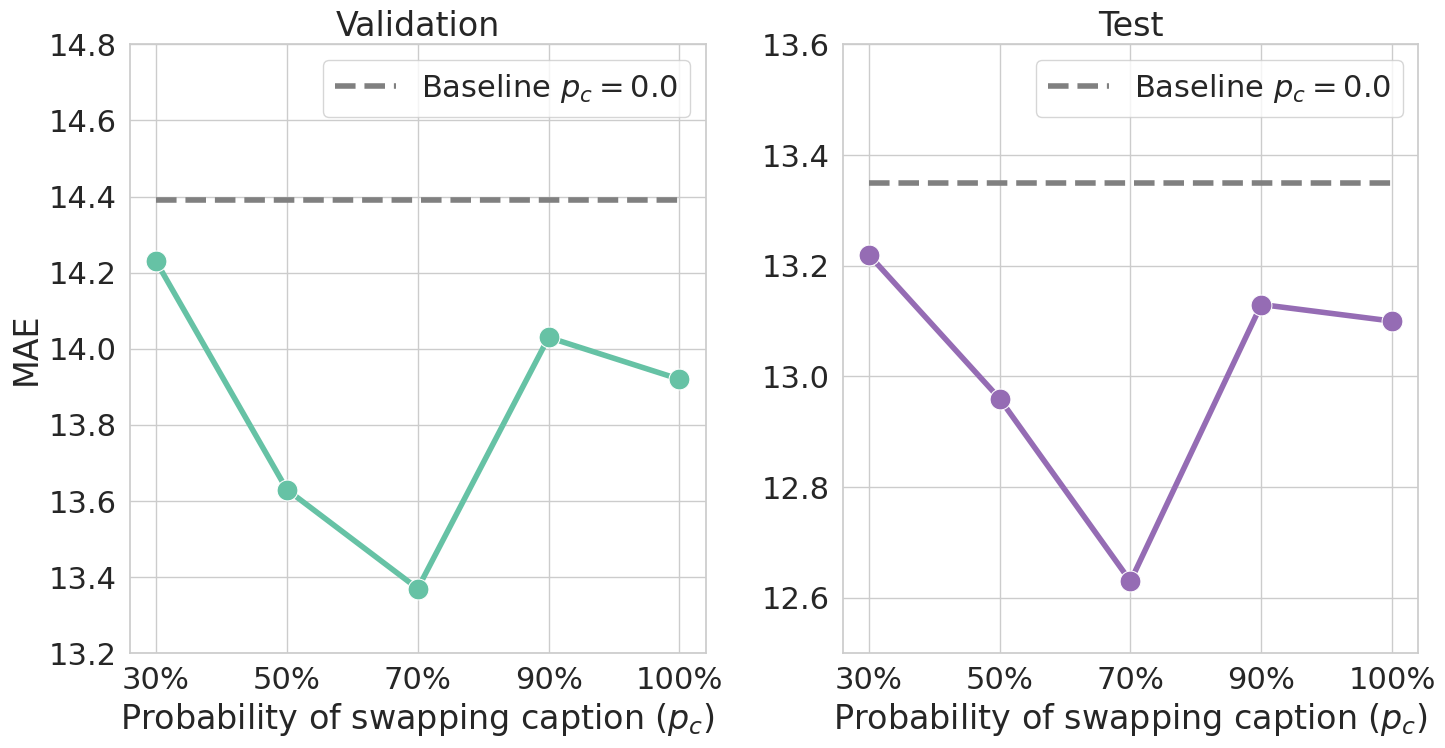}
    \caption{Impact of the percentage of diverse augmentations $p_c$ on SAFECount. MAE is reported for the val and test sets of FSC147.}
    \label{fig:ablation_pc}
    \vspace{-1em}
\end{figure}

\paragraph{Rate of Diverse Samples} In \cref{fig:ablation_pc}, we vary the rate of diverse augmentations among $M=10$ augmentations. We compare with SAFECount trained solely with baseline non-diverse augmentations. More diverse samples overall increase the counting accuracy.
We further find that adding 70\% of diverse samples gives better performances than 50\%.
This suggests that the diverse augmentations are more beneficial than the baseline ones. The augmentations using the original captions yet remain useful to the model as we observe a slight increase in MAE when more than 90\% of the augmentations are obtained on new combinations.%, so it is better to generate them. %thus not generating them at all is not desirable. 

\begin{figure}[t]
    \centering
    \includegraphics[width=8cm,trim=0 5 0 0, clip]{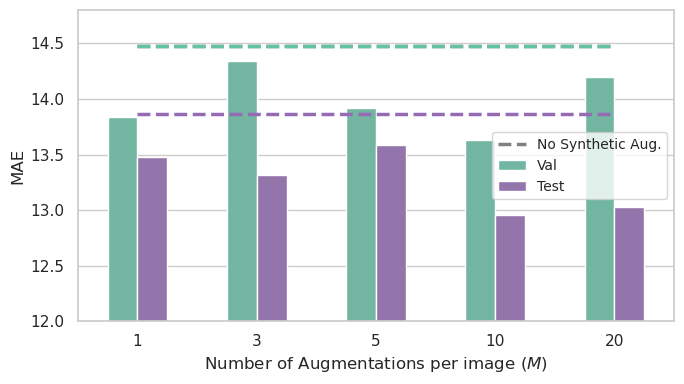}
    \caption{Impact of the number of augmentations $M$ on SAFECount (val. and test MAE on FSC147).}
    \label{fig:ablation_M}
\end{figure}

\paragraph{Number of Augmentations}
In \cref{fig:ablation_M}, we vary the number of augmentations generated for each image. With a single augmentation, the performances already improve. For low values of $M=1,3,5$, the performances are comparable, then a stronger increase is observed for $M=10$. With twice as many augmentations ($M=20$), performances degrade on the validation set. This might be due to an insufficient convergence, as the model is trained with many more different data points but for the same number of iterations.%the model not converging as it sees more data but is trained for the same number of epochs (200).

\begin{figure}
    \centering
    \includegraphics[width=7cm,trim=0 5 0 0, clip]{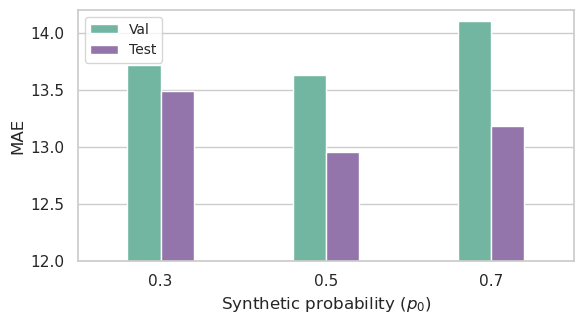}
    \caption{Impact of the synthetic data ratio $p_0$ on SAFECount (val. and test MAE on FSC147).}
    \label{fig:ablation_p0}
    \vspace{-1em}
\end{figure}

\paragraph{Rate of Synthetic Samples} \cref{fig:ablation_p0} shows the counting accuracy w.r.t.\ ratio $p_0$ of synthetic samples vs.\ real samples in a batch. We observe that equally balancing the synthetic and real data gives the best performances, which is consistent with what has been observed in previous works generating synthetic data for image classification \cite{he2022synthetic, bansal2023leaving}. 

\subsection{Generalization on CARPK}

CARPK \cite{hsieh2017drone} was introduced to train networks that can count cars in aerial views of parking lots. It is also used to evaluate the ability of class-agnostic models to count in a class-specific setting. 
Given a model trained on FSC147 (without the examples of the ``car'' category), the model is first evaluated without any fine-tuning, then with fine-tuning on CARPK. We evaluate our model trained on FSC147 with diverse augmentations in the same setting. \cref{tab:quant_results_carpk} reports improved counting performances in both the pre-trained and fine-tuned settings in comparison to the models trained without synthetic augmentations. In the fine-tuning setting, we reach state-of-the-art counting accuracy (4.87 MAE/6.17 RMSE) on CARPK amongst class-agnostic models.

\begin{table}[t]
    \centering
    \begin{tabularx}{\linewidth}{lccc}
    \toprule
     & Aug. & MAE & RMSE \\
    \midrule
    \multirow{2}{*}{\shortstack[l]{Pre-trained on FSC147}} & Trad. %Aug. 
    & 17.65 & 23.83\\
    & Div. %Gen. 
    & \textbf{16.49} & \textbf{19.05} \\
    \midrule
    \multirow{2}{*}{\shortstack[l]{Fine-tuned on CARPK}} & Trad. %Aug. 
    & 5.44 & 6.94 \\
    & Div. %Gen. 
    & \textbf{4.87} & \textbf{6.17} \\
    \midrule
    \midrule
    SAFECount \cite{You_2023_WACV} {\scriptsize(WACV'23)}  & Trad. %Aug. 
    & 5.33 & 7.04 \\
    CounTR \cite{liu2022countr} {\scriptsize(BMVC'22)} & Trad. %Aug.
    & 5.75	& 7.45 \\
    BMNet+ \cite{shi2022represent} {\scriptsize(CVPR'22)} & Trad. %Aug. 
    & 5.76	& 7.83 \\
    \bottomrule
    \end{tabularx}
    \caption{Counting performance on CARPK with SAFECount.}
    \label{tab:quant_results_carpk}
    \vspace{-1em}
\end{table}

%\subsection{Qualitative Results}

\section{Limitation}
\label{sec:limitation}

\begin{figure}[t]
    \centering
    \includegraphics[width=8cm]{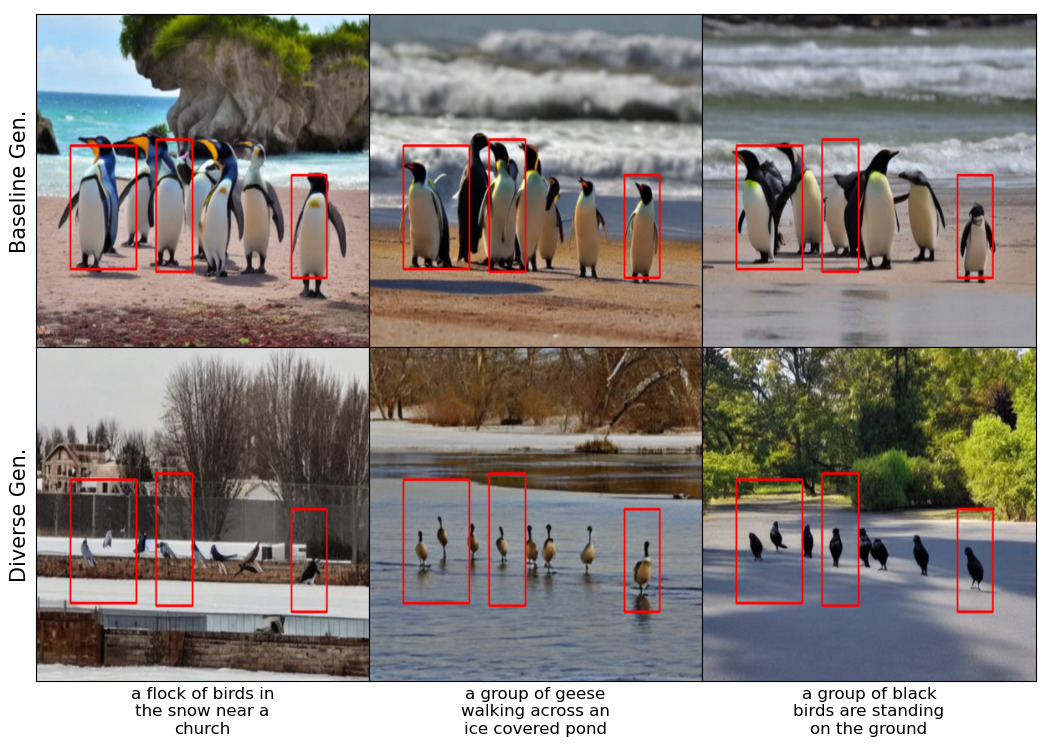}
    \caption{Limitation: our diverse generation strategy can change the size and shape of generated objects, leading to exemplar boxes (in \textcolor{red}{red}) that do not fit perfectly.}
    \label{fig:noisy_exemplars}
    \vspace{-1em}
\end{figure}

Our synthetic data needs a ground truth and exemplars to train the counting network.
Conditioning on densities makes it possible to reuse both the original density and the exemplar bounding boxes. However, changing the caption can affect the object category, and in turn its shape. In some rare cases, exemplar boxes do not fit the generated objects anymore, as illustrated in \cref{fig:noisy_exemplars}. We explored to what extent refining these boxes could improve our model. We segmented objects using SAM in zero-shot \cite{kirillov2023segany} prompted with object centers. Preliminary results showed no improvement with box refinement, possibly due to inaccurate segmentation.

\section{Conclusion}
We show that synthetic data generated by diffusion models improve deep models for few-shot counting. We adapt a pretrained text-to-image model with a density map conditioning and we propose a diversification strategy that exploits caption similarities to generate unseen but plausible data that mixes the semantics and the geometry of different training images. We show that selecting compatible images improves synthetic image quality with beneficial effects on model performance. We demonstrate that learning with our diverse synthetic data leads to improved counting accuracy on FSC147 and state of the art generalization on CARPK. This strategy could be adapted to other tasks requiring fine-grained compositionality, such as object detection and semantic segmentation. Our diversification scheme could be further extended by swapping both the captions and the density controls, by introducing a suitable similarity metric that operates on the density maps.

\textbf{Acknowledgments.} This research was performed under a grant from the AHEAD ANR program (ANR-20-THIA-0002) and supported by the European Commission under European Horizon 2020 Programme, grant number 951911 - AI4Media. This publication was made possible by the use of the FactoryIA supercomputer, financially supported by the Ile-De-France Regional Council. Special thanks to Nicu Sebe and Elia Peruzzo for their advice.

%%%%%%%%% REFERENCES
{\small
\bibliographystyle{ieee_fullname}
\bibliography{egbib}
}

\end{document}